\documentclass[10pt]{article}
\usepackage[margin=0.9in]{geometry}
\usepackage{amsmath,amssymb}
\usepackage{booktabs}
\usepackage{graphicx}
\usepackage{microtype}
\usepackage{url}
\usepackage[hidelinks]{hyperref}
\usepackage[numbers,sort&compress]{natbib}
\usepackage{enumitem}
\usepackage{array}
\usepackage{multirow}
\usepackage{caption}
\usepackage{float}
\usepackage{xspace}
\setlist{nosep,leftmargin=*}
\newcommand{\DeltaAUC}{\Delta_{\mathrm{AUC}}}
\title{\textbf{Not All Data Leakage Is Equal:}\\A Controlled Benchmark of Evaluation Inflation in Tabular Classification}
\author{Ankit Sharma\\International Institute of Information Technology Bangalore\\\texttt{ankitankitsharm.eml36@iiitb.net}}
\date{August 31, 2026}

\begin{document}
\maketitle

\begin{abstract}
Data leakage is widely recognized as a threat to trustworthy machine-learning evaluation, yet the phrase covers mechanisms with very different statistical consequences. We present a controlled empirical benchmark that measures the inflation in held-out ROC-AUC caused by five common leakage mechanisms: global standardization, global median imputation, duplicate contamination, supervised feature selection before splitting, and target encoding before splitting. The benchmark spans six public datasets and two synthetic stress tests, three model families, and five repeated stratified train/test splits, yielding 600 paired clean-versus-leaked comparisons. To avoid pseudo-replication, inference is performed on dataset-level effects with cluster bootstrap confidence intervals and Holm-corrected one-sided Wilcoxon tests. Label-dependent or direct data-reuse mechanisms produced systematic optimism: global target encoding increased ROC-AUC by 2.42 percentage points on average (95\% cluster-bootstrap CI 0.62--4.89), global supervised feature selection by 1.93 points (0.63--3.39), and duplicate contamination by 1.52 points (0.40--2.80). All three remained significant after multiplicity correction. In contrast, global scaling and global median imputation had near-zero average effects under the i.i.d. settings studied here. A controlled target-encoding stress test further shows that leakage becomes sharply worse as categories become sparse. The results support a severity-aware view of leakage: evaluation pipelines should prevent all held-out-data dependence, but label-dependent transformations and observation reuse deserve especially aggressive auditing.
\end{abstract}

\section{Introduction}
Reliable predictive evaluation rests on a simple separation: information used to construct a model must not improperly depend on the examples used to estimate its generalization performance. Violations of this separation are commonly grouped under the term \emph{data leakage}. Leakage has been documented in industrial data mining, biomedical prediction, social-science forecasting, benchmark construction, and machine-learning-based science more broadly \citep{kaufman2012leakage,ambroise2002selection,kapoor2023leakage}. Its practical consequence is often an evaluation score that is too optimistic for deployment.

The label ``leakage,'' however, hides substantial heterogeneity. Fitting a standardization transform on all feature vectors, selecting features using all labels, computing target statistics before a split, and allowing exact test examples into the training set all violate a strict holdout protocol, but they do not expose the learner to the same information. Prior work has developed taxonomies of leakage \citep{kaufman2012leakage,kapoor2023leakage}, analyzed particular preprocessing mechanisms theoretically \citep{moscovich2022cross}, and demonstrated severe selection bias in high-dimensional biomedical settings \citep{ambroise2002selection,varma2006bias}. What is less clear to practitioners is the relative empirical magnitude of several common mistakes when evaluated under one paired protocol.

This paper asks a narrow question:
\begin{quote}
\emph{When the underlying data, train/test split, model family, and evaluation metric are held fixed, how much apparent predictive performance can different leakage mechanisms add?}
\end{quote}

We answer this using a mechanism-isolating benchmark. Every leaked run is paired with a corresponding leakage-safe run on the same split. We focus on tabular binary classification and ROC-AUC because it is threshold independent and widely used for ranking performance \citep{fawcett2006roc}. The benchmark deliberately includes both positive controls expected to cause optimism and label-free preprocessing operations for which effects may be small or sign-indefinite.

Our contributions are:
\begin{enumerate}
    \item \textbf{A common paired benchmark for five leakage mechanisms.} We evaluate global scaling, global imputation, duplicate contamination, global supervised feature selection, and global target encoding across eight datasets and three model families.
    \item \textbf{Severity estimates rather than a binary leakage label.} We report the change in ROC-AUC induced by each mechanism and quantify uncertainty at the dataset level.
    \item \textbf{A mechanism study for target encoding.} Using a deliberately uninformative sparse categorical feature, we show that the optimism from global target encoding increases as category support decreases.
    \item \textbf{Model-specific evidence.} Duplicate contamination is most damaging for random forests in our benchmark, whereas leaked feature selection is more damaging for logistic regression and $k$-nearest neighbors.
\end{enumerate}

The main conclusion is not that label-free preprocessing is universally safe. Indeed, Moscovich and Rosset show that unsupervised preprocessing can bias validation estimates \citep{moscovich2022cross}. Rather, under the i.i.d. tabular regimes and transformations tested here, the largest and most consistent optimism comes from mechanisms that use labels or reuse held-out observations directly.

\section{Related Work}
\paragraph{Definitions and taxonomies.}
Kaufman et al. formalized leakage as the introduction of information about the prediction target that would not legitimately be available at prediction time and discussed both feature- and example-level pathways \citep{kaufman2012leakage}. More recently, Kapoor and Narayanan surveyed machine-learning-based science and documented leakage-related reproducibility failures across many fields, proposing a detailed taxonomy and reporting checklist \citep{kapoor2023leakage}. Our work is complementary: rather than expanding the taxonomy, we place several frequently encountered mechanisms into a single controlled empirical comparison.

\paragraph{Feature selection and model-selection bias.}
The danger of performing supervised selection outside a validation loop is well established. Ambroise and McLachlan showed that selecting genes using all samples before estimating error creates severe selection bias in microarray classification \citep{ambroise2002selection}. Varma and Simon demonstrated that tuning a classifier and reporting the same cross-validation estimate used for selection is optimistic, motivating nested validation \citep{varma2006bias}. Rao et al. experimentally studied overuse of cross-validation in model and feature selection \citep{rao2008dangers}. Our feature-selection intervention follows this lineage but evaluates its effect alongside other leakage mechanisms under the same models and splits.

\paragraph{Unsupervised preprocessing.}
A common intuition is that transformations that do not use labels are harmless before validation. Moscovich and Rosset challenge this intuition theoretically and empirically, showing that preprocessing such as variance filtering, grouping rare categories, and rescaling can bias cross-validation, with sign and magnitude depending on the data-generating and modeling pipeline \citep{moscovich2022cross}. Our results should therefore be interpreted as empirical measurements for the specific i.i.d. regimes studied, not as a general safety theorem.

\paragraph{Target statistics.}
Target encoding maps a category to a statistic of the response. If the statistic for an observation depends on its own label or on held-out labels, target information can enter the feature representation. CatBoost's ordered target statistics were explicitly designed to mitigate such target leakage and prediction shift \citep{prokhorenkova2018catboost}. We use target encoding as a controlled mechanism because the amount of self/held-out influence can be varied through category sparsity.

\section{Problem Setup}
Let $D=\{(x_i,y_i)\}_{i=1}^{n}$ be a labeled dataset and let a split produce disjoint index sets $I_{\mathrm{tr}}$ and $I_{\mathrm{te}}$. A leakage-safe pipeline constructs every learned preprocessing transform and predictor using information permitted by $I_{\mathrm{tr}}$, then applies the resulting functions to $I_{\mathrm{te}}$ without updating them.

For a leakage mechanism $m$, model family $a$, dataset $d$, and random split $s$, let $A^{\mathrm{safe}}_{dasm}$ and $A^{\mathrm{leak}}_{dasm}$ denote held-out ROC-AUC. We define evaluation inflation as
\begin{equation}
(\DeltaAUC)_{dasm} = A^{\mathrm{leak}}_{dasm} - A^{\mathrm{safe}}_{dasm}.
\end{equation}
Positive $\DeltaAUC$ means the leaked pipeline appears better on the holdout than its leakage-safe counterpart.

We distinguish three broad pathways:
\begin{enumerate}
    \item \textbf{Label-free held-out dependence}: a transform uses $x$ values from the test set but not $y$ (global scaling and imputation).
    \item \textbf{Label-dependent held-out dependence}: a transform uses test labels, directly or through a supervised statistic (global feature selection and global target encoding).
    \item \textbf{Observation reuse}: test examples or their exact duplicates enter training (duplicate contamination).
\end{enumerate}
This partition is descriptive rather than exhaustive; temporal leakage, entity leakage, label-proxy features, post-outcome measurements, and pretraining contamination are outside the present benchmark.

\section{Benchmark Design}
\subsection{Datasets}
We use six public datasets available locally through scikit-learn \citep{pedregosa2011scikit} or statsmodels \citep{seabold2010statsmodels}, plus two synthetic stress tests generated with \texttt{make\_classification}. Multi-class public datasets are converted to deterministic binary tasks so that the same metric and models can be used throughout. Table~\ref{tab:datasets} summarizes the resulting tasks. The Digits data are derived from the UCI optical-recognition dataset \citep{alpaydin1998digits}; Iris originates with Fisher's classical study \citep{fisher1936taxonomic}; the Affairs data originate with Fair \citep{fair1978theory}; and ANES96 is a public subset of the 1996 American National Election Study \citep{anes1996}. The Breast Cancer and Wine datasets are standard UCI-derived datasets distributed by scikit-learn \citep{uci2019,pedregosa2011scikit}.

\begin{table}[t]
\centering
\caption{Benchmark datasets. Positive rate is the fraction of class-1 observations after binary task construction. Clean AUC is averaged over the three model families and five splits using the ordinary, uncontaminated pipeline.}
\label{tab:datasets}
\small
\begin{tabular}{lrrrr}
\toprule
Dataset & $n$ & $p$ & Positive rate & Clean AUC \\
\midrule
Breast Cancer & 569 & 30 & 0.627 & 0.985 \\
Digits $\geq 5$ & 1,797 & 64 & 0.499 & 0.983 \\
Wine class 0 & 178 & 13 & 0.331 & 0.997 \\
Iris virginica & 150 & 4 & 0.333 & 0.991 \\
Fair Affairs & 6,366 & 8 & 0.322 & 0.703 \\
ANES96 vote & 944 & 10 & 0.416 & 0.957 \\
Synthetic Small & 250 & 20 & 0.500 & 0.849 \\
Synthetic High-Dim & 800 & 150 & 0.502 & 0.735 \\
\bottomrule
\end{tabular}
\end{table}

\subsection{Models and evaluation protocol}
We select three model families with different inductive biases:
\begin{itemize}
    \item logistic regression with $\ell_2$ regularization;
    \item 5-nearest-neighbor classification \citep{cover1967nearest};
    \item a random forest with 50 trees \citep{breiman2001random}.
\end{itemize}
No hyperparameter search is performed, because tuning would add another selection layer to the experiment. Logistic regression and $k$-NN receive standardization learned from the training data except in the scaling intervention, where the scaler itself is the treatment. Random forests are included as a scale-invariant negative control in the scaling experiment.

For each dataset we generate five stratified 75/25 train/test splits with seeds 0--4. Every leakage intervention is paired to a clean pipeline on exactly the same split and model. The main benchmark therefore contains
\[
8\ \text{datasets}\times 5\ \text{leakage mechanisms}\times 3\ \text{models}\times 5\ \text{splits}=600
\]
paired comparisons. We compute ROC-AUC, log loss, and accuracy; ROC-AUC is the primary endpoint.

\subsection{Leakage interventions}
\paragraph{Global scaling.}
The safe pipeline fits a standard scaler only on $X_{\mathrm{tr}}$. The leaked pipeline fits mean and variance parameters on the full feature matrix $X$ before applying the split. No labels are used.

\paragraph{Global imputation.}
We inject 20\% missing-completely-at-random entries into each feature matrix using a fixed mask per split. The safe pipeline learns column medians from the training partition. The leaked pipeline computes medians from the entire feature matrix before splitting. Again, no labels are used.

\paragraph{Duplicate contamination.}
The safe model is trained on the ordinary training set. For the leaked condition, exact copies of a randomly selected 20\% of test observations, including their labels, are appended to the training set. The test set itself is unchanged. This intervention isolates observation reuse while holding the evaluation population fixed.

\paragraph{Global supervised feature selection.}
To create a setting where selection noise can be measured across heterogeneous datasets, we augment each dataset with independent Gaussian noise features. The number of added features is $\min(300,\max(150,4p))$. Both pipelines select the best $k=\min(25,p_{\mathrm{aug}})$ features using the univariate ANOVA $F$ statistic. In the safe pipeline, selection is fit on training data only. In the leaked pipeline, the selector is fit once using all feature vectors and all labels before the split. This is a semi-synthetic stress test of selection bias rather than a claim about the original feature spaces.

\paragraph{Global target encoding.}
We append a deliberately uninformative categorical variable independent of $X$ and $y$, with approximately five observations per category. The safe pipeline forms leave-one-out smoothed target encodings for training rows and uses training-only category statistics for test rows. The leaked pipeline computes smoothed category target means using the full dataset, including test labels, before the split. Because the categorical variable has no population-level signal, any systematic increase in held-out performance is attributable to the encoding pathway rather than a useful underlying feature.

\subsection{Statistical analysis}
Seeds and models from the same dataset are not independent scientific replications. We therefore first average $\DeltaAUC$ over the five splits and three models within each dataset, leaving eight dataset-level effects per leakage mechanism. We report their mean and a 95\% nonparametric cluster-bootstrap confidence interval obtained by resampling datasets with replacement 50,000 times. To test whether a mechanism produces positive inflation, we use a one-sided Wilcoxon signed-rank test across the eight dataset-level effects. The five resulting $p$-values are adjusted using Holm's sequential procedure \citep{holm1979simple}. These tests are descriptive evidence over the benchmark collection, not population-level guarantees over all tabular tasks.

\section{Results}
\subsection{Label-dependent leakage and observation reuse produce systematic optimism}
\begin{figure}[t]
\centering
\includegraphics[width=0.95\linewidth]{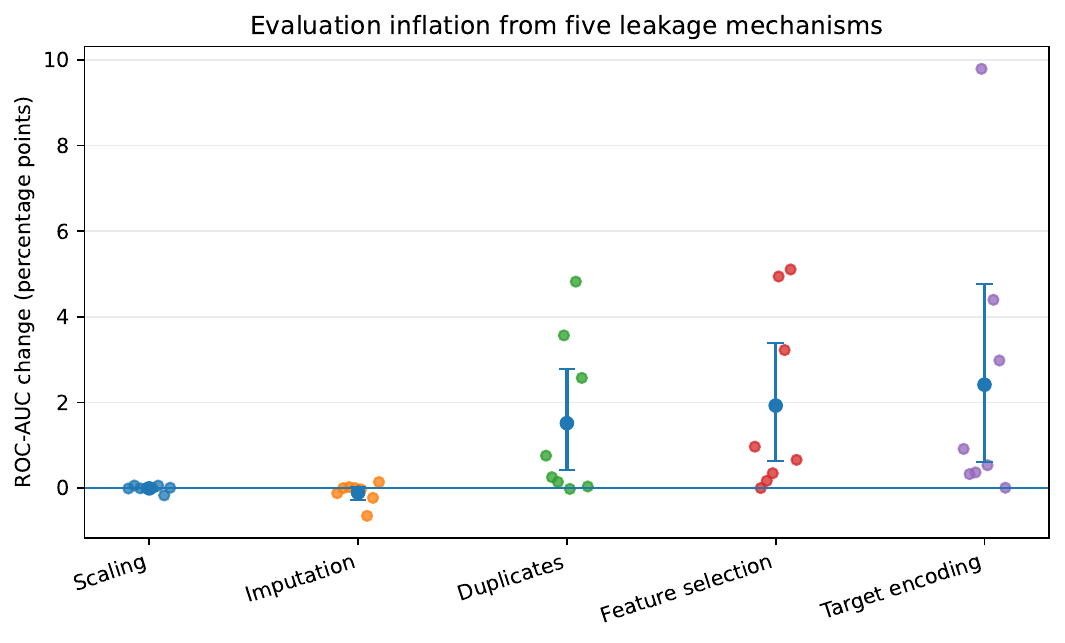}
\caption{Dataset-level ROC-AUC inflation for five leakage mechanisms. Small markers show the eight dataset-level effects, each already averaged over three model families and five splits. Large markers show the mean; error bars are 95\% cluster-bootstrap confidence intervals.}
\label{fig:main}
\end{figure}

Figure~\ref{fig:main} and Table~\ref{tab:main} show the primary result. Global target encoding produces the largest mean inflation, +2.42 ROC-AUC percentage points (95\% cluster-bootstrap CI 0.62--4.89). Global supervised feature selection adds +1.93 points (0.63--3.39), and duplicate contamination adds +1.52 points (0.40--2.80). All three effects remain significant after Holm correction at the 0.05 level.

In contrast, global scaling has a mean effect of $-0.008$ points (95\% CI $-0.059$ to $0.029$) and global median imputation has a mean effect of $-0.108$ points ($-0.284$ to $0.023$). Neither is significant. These near-zero averages do not contradict prior results showing that unsupervised preprocessing can bias validation \citep{moscovich2022cross}; rather, they indicate that for these particular transformations, datasets, split sizes, and model choices, the net empirical impact is small.

\begin{table}[t]
\centering
\caption{Primary inference at the dataset level. Effects are ROC-AUC percentage points. Holm-adjusted $p$-values come from one-sided Wilcoxon signed-rank tests over eight dataset-level effects.}
\label{tab:main}
\small
\begin{tabular}{lrrrr}
\toprule
Mechanism & Mean & 95\% CI & Positive datasets & Holm $p$ \\
\midrule
Global target encoding & 2.42 & [0.62, 4.89] & 8/8 & 0.0195 \\
Global feature selection & 1.93 & [0.63, 3.39] & 7/8 & 0.0313 \\
Duplicate contamination & 1.52 & [0.40, 2.80] & 7/8 & 0.0313 \\
Global scaling & -0.01 & [-0.06, 0.03] & 4/8 & 0.9375 \\
Global imputation & -0.11 & [-0.28, 0.02] & 2/8 & 0.9375 \\
\bottomrule
\end{tabular}
\end{table}

The secondary metrics point in the same direction. Averaged over all seed/model/dataset pairs, the three high-impact mechanisms improve apparent accuracy by 1.18--2.21 percentage points and reduce apparent log loss by 0.055--0.087. Thus, the main result is not an artifact of the ranking-only AUC metric (Appendix~\ref{app:secondary}).

\subsection{Leakage severity depends on the learner}
\begin{figure}[t]
\centering
\includegraphics[width=0.88\linewidth]{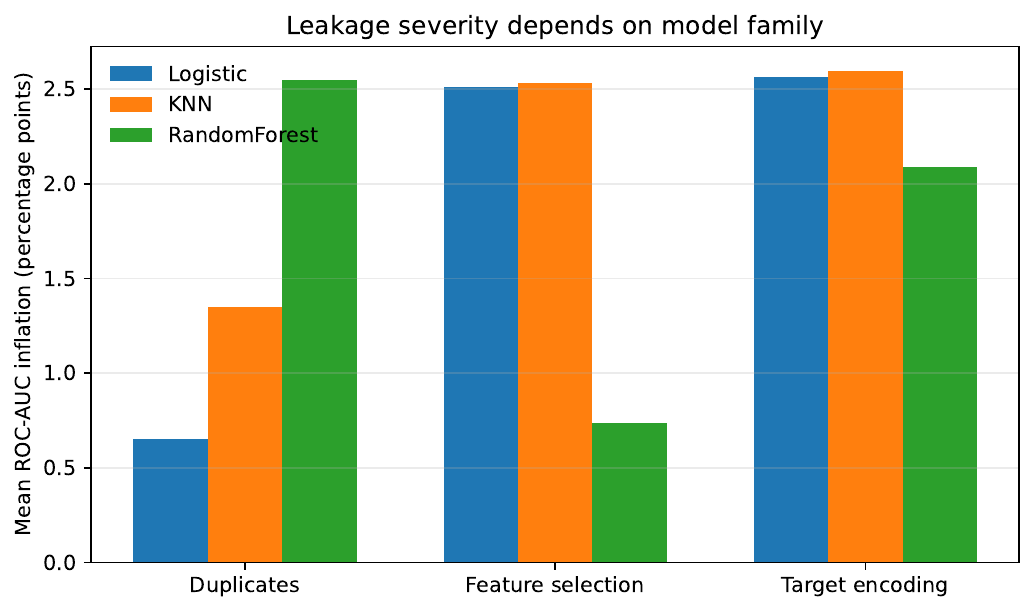}
\caption{Mean ROC-AUC inflation by model family for the three mechanisms with systematic optimism. Means are over eight datasets and five splits.}
\label{fig:model}
\end{figure}

The same contamination mechanism does not affect all learners equally (Figure~\ref{fig:model}). Random forests show the largest mean AUC inflation from exact duplicate contamination (+2.55 points), consistent with their ability to isolate highly specific regions of feature space. Logistic regression is less affected (+0.65 points). In contrast, global feature selection inflates logistic regression and $k$-NN by about +2.5 points but random forests by only +0.74 points, plausibly because the forest can down-weight many irrelevant features internally. Target encoding affects all three families comparably (+2.09 to +2.60 points).

These interactions argue against attaching a universal ``leakage penalty'' to a preprocessing mistake. The severity of contamination depends on how strongly a learner can exploit the leaked signal.

\subsection{Sparse categories amplify target-encoding leakage}
\begin{figure}[t]
\centering
\includegraphics[width=0.86\linewidth]{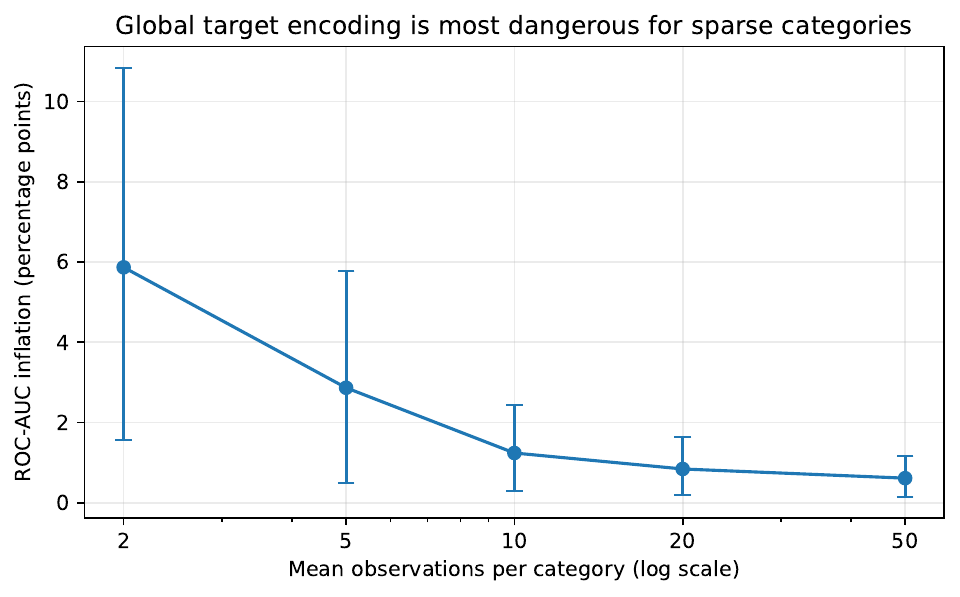}
\caption{Target-encoding sensitivity for logistic regression. Points are means over eight dataset-level effects; error bars are 95\% cluster-bootstrap confidence intervals. The x-axis is logarithmic. Sparse categories allow an individual held-out label to contribute a larger fraction of its encoded feature value.}
\label{fig:te}
\end{figure}

To test the mechanism behind global target encoding, we repeat the target-encoding experiment for logistic regression while varying the expected number of observations per random category from 2 to 50. The categorical feature remains independent of the target by construction. Figure~\ref{fig:te} shows a clear severity gradient: mean AUC inflation is 5.87 points when categories contain about two observations, 2.87 points at five observations, 1.24 at ten, 0.84 at twenty, and 0.61 at fifty.

The mechanism is straightforward. If a category contains $c$ observations and a global target mean is computed before splitting, the label of a held-out row can directly influence the encoding assigned to that row with weight on the order of $1/c$ (modulo smoothing). As $c$ grows, the self/held-out contribution is diluted. This controlled result is consistent with the motivation for ordered or out-of-fold target statistics \citep{prokhorenkova2018catboost}.

\subsection{Effects vary sharply across datasets}
Average effects hide substantial heterogeneity. For example, target-encoding leakage inflates AUC by 9.79 points on Fair Affairs and 4.40 points on Synthetic High-Dim, but is effectively zero on Wine class 0, whose clean AUC is already near 1.0. Feature-selection leakage is largest on Iris (+4.94 points) and Synthetic Small (+5.11), while it is negligible on Breast Cancer in this configuration. Duplicate contamination is most visible on Synthetic High-Dim (+4.82) and Fair Affairs (+3.57). Full dataset-level results appear in Appendix~\ref{app:dataset}.

Ceiling effects are one reason for this heterogeneity: when clean AUC is already 0.99--1.00, little upward movement is possible. More generally, sample size, dimensionality, class overlap, model fit, and the chance structure induced by the intervention all affect how much leaked information a learner can exploit.

\section{Discussion}
\subsection{Leakage should be audited by information pathway}
A practical takeaway is that ``did preprocessing happen before the split?'' is necessary but not sufficiently diagnostic. The more informative audit question is: \emph{what information from the held-out set can influence the fitted representation or model, and how directly?}

Our benchmark suggests the following qualitative ordering for common tabular workflows:
\begin{enumerate}
    \item \textbf{Highest concern: label-dependent transforms.} Feature selection, target encoding, supervised dimensionality reduction, and threshold tuning can search over held-out labels and convert chance correlations into apparent signal.
    \item \textbf{Highest concern: observation reuse.} Duplicate or near-duplicate records crossing a split can turn generalization evaluation into partial memorization, especially for flexible local or tree-based models.
    \item \textbf{Still invalid, but often lower empirical impact in i.i.d. settings: label-free global transforms.} Scaling or simple imputation can use held-out covariates and therefore violate a strict validation protocol, but their average effect in our experiments is small. This should not be generalized to all unsupervised transformations or distribution-shift settings.
\end{enumerate}

The safest engineering rule remains simple: \textbf{split first, then fit every data-dependent operation inside the training partition or validation fold}. Pipeline abstractions make this easier to enforce. Nested validation is required when feature selection, model selection, or hyperparameter tuning is itself optimized \citep{varma2006bias}.

\subsection{Why negative results matter}
A benchmark designed only around spectacular failures risks giving practitioners the wrong intuition about prevalence and magnitude. The near-zero scaling and imputation effects are useful negative controls. They show that the mere existence of held-out covariate dependence does not imply large positive AUC inflation in every finite sample. This is consistent with the theoretical point that unsupervised-preprocessing bias can be positive, negative, or small depending on the full pipeline \citep{moscovich2022cross}.

\subsection{Benchmarking versus real-world leakage}
Our interventions are intentionally controlled. Real leakage is often harder: patient images may share a subject across folds, future information may be backfilled into historical rows, labels may be encoded in operational fields, or benchmark examples may enter pretraining corpora. Such cases can combine several mechanisms and create distribution-dependent effects far larger than those reported here. The value of a controlled benchmark is not realism in every detail, but causal interpretability of the comparison.

\section{Limitations}
This study has several limitations. First, it is restricted to binary tabular classification and three classical model families; neural networks, gradient boosting systems with specialized categorical handling, regression, ranking, survival analysis, and structured prediction may respond differently. Second, the feature-selection and target-encoding interventions are semi-synthetic by design: noise features and an uninformative categorical variable are injected to isolate mechanisms across heterogeneous datasets. Third, only five train/test splits are used; we mitigate overinterpretation by treating datasets as the independent units, but larger benchmark collections would support narrower confidence intervals. Fourth, the benchmark assumes random i.i.d. splits and does not study temporal, grouped, geographic, or entity-disjoint evaluation. Fifth, several public tasks are easy and exhibit AUC ceiling effects. Finally, this work measures evaluation inflation, not downstream harm; the operational cost of a one-point AUC error is application dependent.

These limitations define natural extensions: larger OpenML-scale benchmarks, temporal and entity leakage, near-duplicate rather than exact-duplicate contamination, regression and calibration metrics, and automated leakage detectors.

\section{Reproducibility and Data Availability}
The arXiv source bundle accompanying this manuscript contains the benchmark scripts, target-encoding sensitivity analysis, plotting code, raw paired results, and aggregate statistical summaries. The real datasets used here are distributed with scikit-learn or statsmodels, so the experiments do not require downloading private or proprietary data. Synthetic datasets are generated deterministically from fixed seeds.

The recorded environment is Python 3.13.5, NumPy 2.3.5, pandas 2.2.3, SciPy 1.17.0, scikit-learn 1.8.0, statsmodels 0.14.6, and Matplotlib 3.10.8. Random seeds for the primary train/test splits are 0--4. The exact scripts define all additional seeds used for missingness masks, noise features, categorical variables, and duplicate selection.

\section{Conclusion}
Data leakage is not a single empirical phenomenon. Across eight tabular classification tasks, three label-dependent or direct-reuse mechanisms---global target encoding, global supervised feature selection, and duplicate contamination---produce consistent and statistically detectable optimism, while global standardization and global median imputation have near-zero average effects in the i.i.d. regimes studied. The strongest controlled mechanism is sparse target encoding: when categories contain few observations, a held-out label can materially influence its own representation.

The practical recommendation is conservative but severity-aware. All learned preprocessing should remain inside the training or validation fold, because even label-free transformations can bias evaluation. At the same time, auditing effort should prioritize transformations that consume labels, operations that select among many candidate representations, and any process that can place the same entity or observation on both sides of an evaluation boundary.

\appendix
\section{Full Dataset-Level Effects}
\label{app:dataset}
Table~\ref{tab:dataset_effects} reports mean ROC-AUC changes in percentage points after averaging over five splits and three model families.

\begin{table}[H]
\centering
\caption{Dataset-level mean ROC-AUC change (percentage points). Positive values indicate optimistic evaluation.}
\label{tab:dataset_effects}
\scriptsize
\begin{tabular}{lrrrrr}
\toprule
Dataset & Scaling & Imputation & Duplicates & Feature sel. & Target enc. \\
\midrule
ANES96 Vote & -0.01 & -0.12 & 0.76 & 0.97 & 0.91 \\
Breast Cancer & 0.06 & -0.00 & 0.25 & 0.00 & 0.33 \\
Digits $\geq 5$ & -0.01 & 0.02 & 0.14 & 0.17 & 0.37 \\
Fair Affairs & 0.00 & -0.00 & 3.57 & 0.35 & 9.79 \\
Iris virginica & 0.00 & -0.03 & -0.02 & 4.94 & 0.53 \\
Synthetic High-Dim & 0.06 & -0.65 & 4.82 & 3.22 & 4.40 \\
Synthetic Small & -0.17 & -0.23 & 2.57 & 5.11 & 2.98 \\
Wine class 0 & 0.01 & 0.14 & 0.04 & 0.66 & 0.01 \\
\bottomrule
\end{tabular}
\end{table}

\section{Secondary Metrics}
\label{app:secondary}
Table~\ref{tab:secondary} reports averages over all dataset/model/split pairs. For log loss, negative change means the leaked pipeline appears better.

\begin{table}[H]
\centering
\caption{Secondary metric changes averaged over all 120 paired runs per mechanism. Accuracy values are percentage-point changes; log-loss values are raw changes.}
\label{tab:secondary}
\small
\begin{tabular}{lrr}
\toprule
Mechanism & Accuracy change (pp) & Log-loss change \\
\midrule
Global target encoding & 2.21 & -0.0749 \\
Global feature selection & 1.85 & -0.0869 \\
Duplicate contamination & 1.18 & -0.0550 \\
Global scaling & -0.02 & 0.0039 \\
Global imputation & -0.07 & -0.0082 \\
\bottomrule
\end{tabular}
\end{table}

\section{Model-Family Effects}
\begin{table}[H]
\centering
\caption{Mean ROC-AUC inflation (percentage points) by model family for the three high-impact leakage mechanisms.}
\small
\begin{tabular}{lrrr}
\toprule
Mechanism & Logistic & $k$-NN & Random forest \\
\midrule
Duplicate contamination & 0.65 & 1.35 & 2.55 \\
Global feature selection & 2.51 & 2.53 & 0.74 \\
Global target encoding & 2.56 & 2.60 & 2.09 \\
\bottomrule
\end{tabular}
\end{table}

\section{Target-Encoding Sensitivity Values}
\begin{table}[H]
\centering
\caption{Mean ROC-AUC inflation for logistic regression as target-encoding category support changes. Values are averaged over eight datasets and five splits.}
\small
\begin{tabular}{rr}
\toprule
Mean observations per category & AUC inflation (pp) \\
\midrule
2 & 5.87 \\
5 & 2.87 \\
10 & 1.24 \\
20 & 0.84 \\
50 & 0.61 \\
\bottomrule
\end{tabular}
\end{table}

\section{Implementation Notes}
The main code is provided in \texttt{code/main\_benchmark.py}. The leakage-safe target-encoding implementation and corrected main target-encoding runs are in \texttt{code/target\_encoding\_safe.py}; the density sensitivity experiment is in \texttt{code/target\_encoding\_sensitivity.py}. Result files are stored under \texttt{results/}. We intentionally retain seed-level results rather than only aggregates so that alternative hierarchical analyses can be performed without rerunning the models.

\end{document}